\documentclass{article} 
\usepackage[preprint]{neurips_2023}
\usepackage{natbib}

\usepackage[dvipsnames]{xcolor}

\definecolor{cvprblue}{rgb}{0.21,0.49,0.74}
\usepackage[pagebackref,breaklinks,colorlinks,citecolor=cvprblue]{hyperref}
\usepackage{pifont}
\usepackage{amsmath}
\usepackage{algorithm}
\usepackage{algpseudocode}
\usepackage{booktabs}
\usepackage{multirow}
\usepackage{makecell}
\usepackage{colortbl}
\usepackage{graphicx}
\usepackage{wrapfig}
\usepackage{subfigure}

\title{Modality Plug-and-Play: Elastic Modality Adaptation in Multimodal LLMs for Embodied AI}

\author{%
		Kai Huang, Boyuan Yang and Wei Gao \\
		University of Pittsburgh\\
	\texttt{k.huang, by.yang, weigao@pitt.edu} \\
}

\begin{document}
\maketitle

\begin{abstract}
	Large Language Models (LLMs) are capable of reasoning over diverse input data modalities through pre-trained encoders. However, 
	the growing diversity of input data modalities prevents incorporating all modalities into LLMs, especially when LLMs are deployed on resource-constrained edge devices for embodied AI applications. Instead, a better option is to adaptively involve only the useful modalities at runtime, depending on the current environmental contexts and task requirements. For such modality adaptation, existing work adopts fixed connections between encoders and the LLM's input layer, leading to high training cost at runtime and ineffective cross-modal interaction. In this paper, we address these limitations by presenting mPnP-LLM, a new technique that allows fully elastic, automated and prompt runtime modality adaptation, by connecting unimodal encoders to a flexible set of last LLM blocks and making such latent connections fully trainable at runtime. Experiments over the nuScenes-QA dataset show that mPnP-LLM can achieve up to 3.7$\times$ FLOPs reduction and 30\% GPU memory usage reduction, while retaining on-par accuracy with the existing schemes. Under the same compute budget, mPnP-LLM improves the task accuracy by up to 4\% compared to the best existing scheme. Source codes of mPnP-LLM can be found at \url{https://github.com/pittisl/mPnP-LLM}.
\end{abstract}

\vspace{-0.15in}
\section{Introduction}
\vspace{-0.05in}
\label{sec:intro}
Large Language Models (LLMs) can do reasoning over diverse input data modalities [\citenum{alayrac2022flamingo, driess2023palm, brohan2023rt, moon2023anymal}], besides the natural language domain. Such multimodal reasoning relies on pre-trained encoders for different input modalities, such as RGB frames [\citenum{he2022masked}], LiDAR point clouds [\citenum{hess2023masked}] and acoustic signals [\citenum{li2023ti}], to extract task-relevant features. By incorporating encoders from multiple input modalities, LLMs can fully perceive the physical world and enable intelligent embodied agents, such as autonomous vehicles and robots, that adapt to environmental contexts and task needs [\citenum{huang2022inner, driess2023palm, song2023llm}].

\begin{figure}
	\centering
	\vspace{-0.25in}
	\includegraphics[width=4in]{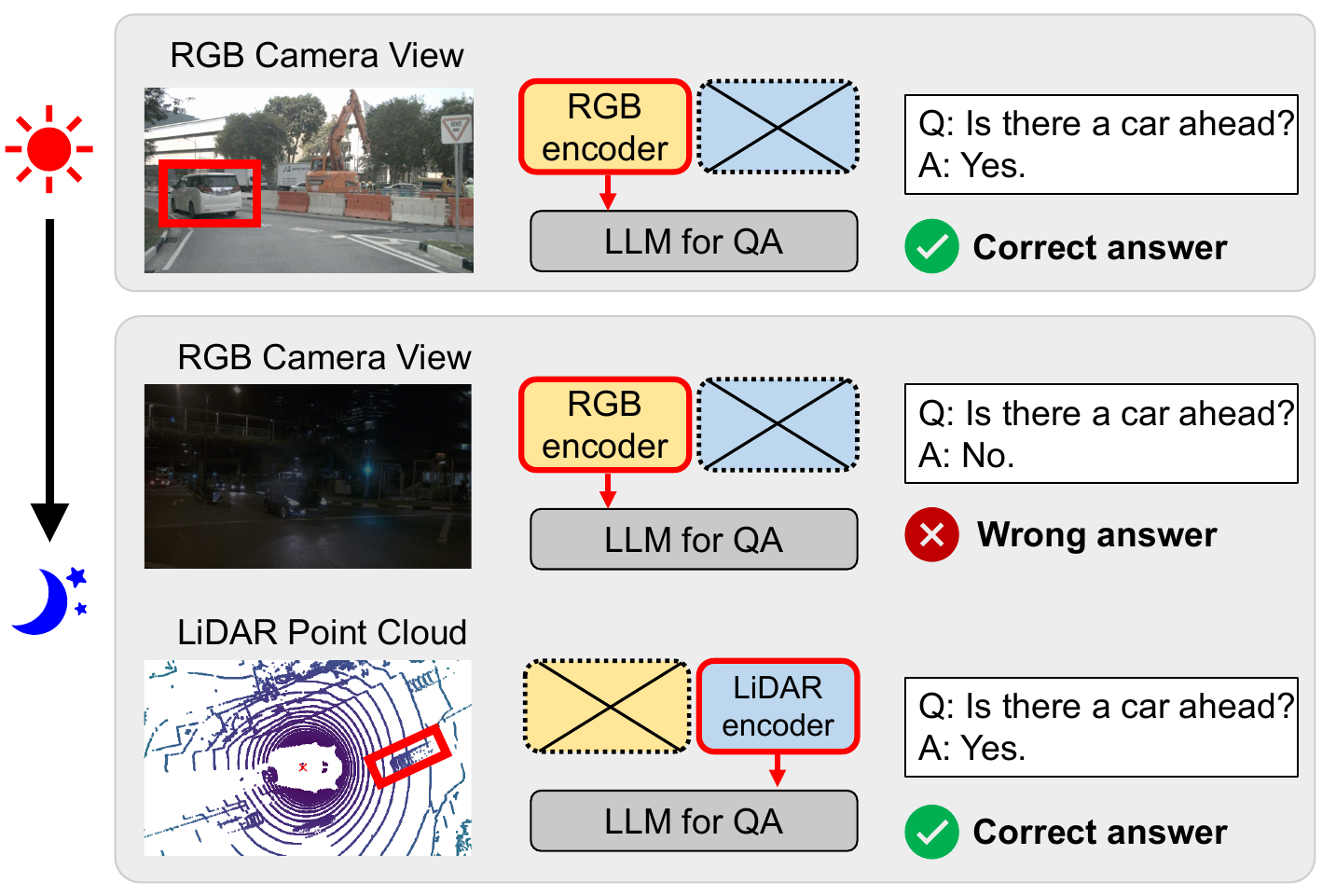}
	\vspace{-0.05in}
	\caption{Runtime modality adaptation for multimodal QA in autonomous driving: During daytime with high visibility, RGB encoder can sufficiently enable visual perception for LLM. However, it fails to provide useful information in nighttime with low visibility. In these cases, we switch to the LiDAR encoder to provide distance perception and retain QA accuracy.}		
	\vspace{-0.15in}
	\label{fig:modality_adaptation}
\end{figure}

One major challenge of such multimodal reasoning is the growing diversity of input data modalities [\citenum{kong2022m3track, xu2022mask, jiao2023bioscatter, fan2023apg}]. Since today's transformer-based encoders [\citenum{vaswani2017attention,wu2021autoformer}] are usually large in size\footnote{For example, the parameter sizes of widely used vision transformers (ViT) can vary between 86M to 22B [\citenum{dosovitskiy2020image}].}, incorporating all modalities to LLMs is computationally expensive or even infeasible, especially on resource-constrained edge devices used in embodied AI applications\footnote{Most hardware platforms used in embodied AI have much weaker capabilities in computation and storage. For example, Nvidia JetBots [\citenum{jetbot}] and Skydio drones [\citenum{skydio}] both have $<$8GB memory, and their GPU computing power is $<$7 TOPS which is $<$1\% of workstation-grade GPUs.}. Instead, we envision that the usefulness of different input modalities, even for the same embodied AI task, could greatly vary when the environmental contexts change. Hence, a better option is to adaptively involve only the useful modalities at runtime, for the minimum on-device computing cost. An example of such runtime modality adaptation for multimodal question answering (QA) tasks in autonomous driving is shown in Figure \ref{fig:modality_adaptation}.


\begin{wrapfigure}{r}{2.5in}
	\centering
	\vspace{-0.1in}
	\includegraphics[width=2.5in]{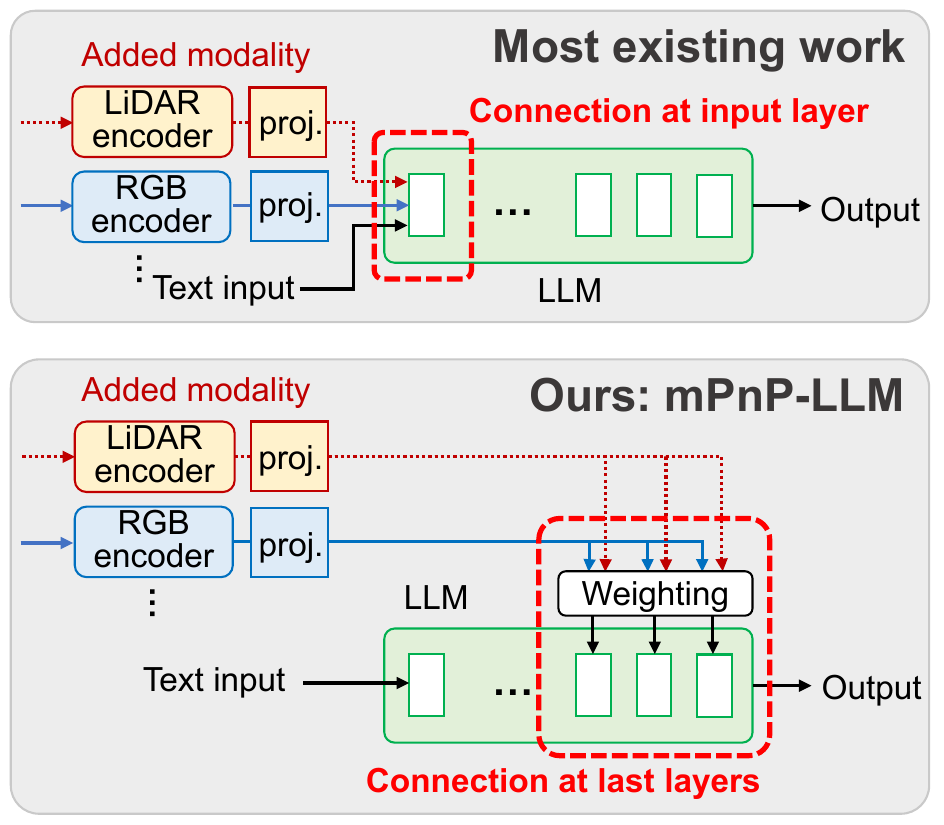}
	\vspace{-0.1in}
	\caption{Existing work vs. mPnP-LLM for modality adaptation, where encoders are connected to LLM via trainable projectors.}
	\vspace{-0.1in}
	\label{fig:snapshot}
\end{wrapfigure}

An intuitive approach to such modality adaptation is to jointly train the encoders of all involved modalities with LLM to align every modality with the natural language domain [\citenum{brohan2023rt, driess2023palm, wu2023next}], but is too expensive for runtime. Instead, we can freeze both encoders and LLM at runtime, but only train the inserted projection modules in between. As shown in Figure \ref{fig:snapshot} - Top, existing work connects encoders to LLM's \emph{input layer} through a trainable projector [\citenum{li2023blip, zhu2023minigpt}], and then applies parameter-efficient LLM fine-tuning [\citenum{hu2021lora, sung2022vl, liang2022modular}] to improve accuracy. However, they still require backpropagating activation gradients throughout the entire LLM and hence incur large training costs. 

Adopting more lightweight projectors or reducing the amount of trainable parameters can only reduce the FLOPs of computing weight updates in training, but do not change the cost of gradient backpropagation. In addition, this connection also requires the projected multimodal features to be all aligned with the LLM input layer's text embedding, but such alignment can be inefficient in bridging the semantic gap across different modalities, because the text representation in early LLM blocks can be too superficial to match the details in input modalities [\citenum{snoek2005early}]. Although some recent work [\citenum{shukor2023ep}] explored connections to LLM's \emph{intermediate layers}, such connections are arbitrarily decided and always fixed at runtime, and hence lack runtime adaptability. Improperly inserting multimodal information to intermediate LLM blocks could also interfere with the LLM's reasoning and affect efficient cross-modal interaction.





To address these limitations, in this paper we present Modality Plug-and-Play in multimodal LLMs (mPnP-LLM), a new technique for  elastic, automated and prompt runtime modality adaptation in multimodal LLMs, by connecting unimodal encoders to a flexible set of last LLM blocks and making such latent connections fully trainable at runtime. As shown in Figure \ref{fig:snapshot} - Bottom, we can adaptively adjust the amount of LLM blocks being connected for different tradeoffs between accuracy and runtime training cost. We can also optimize the efficiency of cross-modal interaction and hence improve accuracy, by controlling the amount of information being injected in each connection with a trainable weighting module. Our design focuses on decoder-only LLM, which is the dominant LLM architecture due to stronger generative power [\citenum{chowdhery2022palm}] and has been widely adopted by most existing LLMs, ranging from BLOOM-1.1B [\citenum{muennighoff2022crosslingual}] and OPT-1.3B [\citenum{zhang2022opt}] to GPT3-175B [\citenum{brown2020language}]. 

Being different from the existing plug-and-play approaches [\citenum{de2017modulating, tan2019lxmert, brohan2022rt, shukor2023ep}] that  require re-implementation of LLM's source codes, mPnP-LLM inserts the projected multimodal tokens as new key-value pairs into the multi-head attention (MHA) module of LLM block. Since such interfacing is well supported in popular LLM frameworks (e.g., HuggingFace Transformers [\citenum{wolf2019huggingface}]), we can avoid any manual programming and reconfiguration efforts at runtime.

We implemented and evaluated mPnP-LLM with two open-sourced LLMs, namely OPT [\citenum{zhang2022opt}] and BLOOMZ [\citenum{muennighoff2022crosslingual}], on the nuScene-QA dataset [\citenum{qian2023nuscenes}] for multimodal QA task in autonomous driving. When adapting between modalities at runtime, mPnP-LLM achieves up to 3.7$\times$ FLOPs reduction and 30\% GPU memory usage reduction, while retaining on-par accuracy with the existing schemes. Such speedup enables modality adaptation within a few minutes on a weak edge device (NVidia Jetson AGX Orin). Under the same compute budget, mPnP-LLM improves the task accuracy by up to 4\%, compared to the best existing scheme.

\section{Background, Existing Work \& Motivation}
\label{sec:background_and_motivation}
To motivate our design of mPnP-LLM, we first describe how the LLM architecture exhibits opportunities of receiving additional inputs for modality adaptation, and then describe the inefficiency of existing schemes to inject multimodal information. Their limitations drive us to enforce elasticity in modality adaptation and look into the backpropagation model for runtime training acceleration.


\begin{wrapfigure}{R}{3in}
	\centering
	\vspace{-0.2in}
	\includegraphics[width=3in]{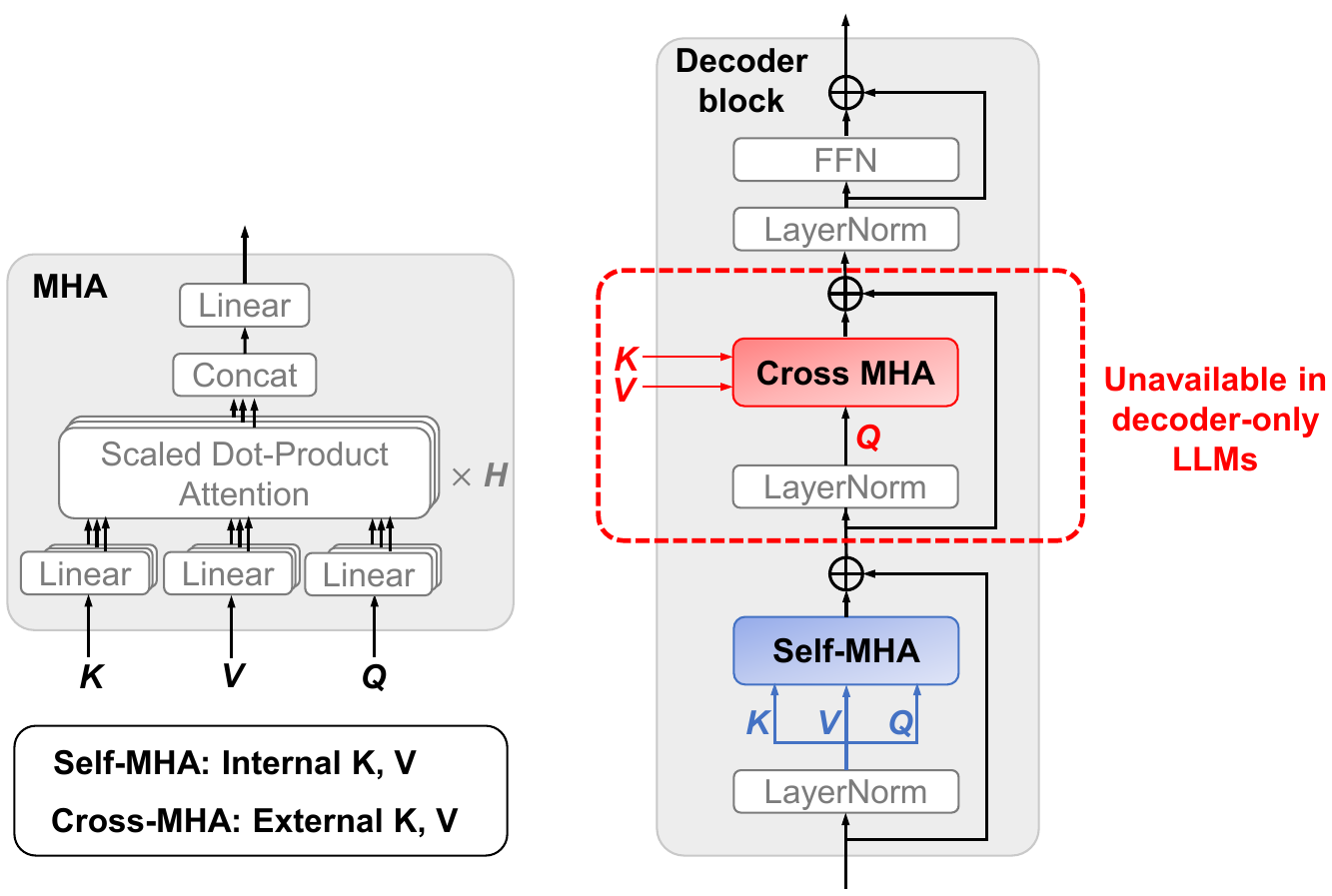}
	\vspace{-0.15in}
	\caption{LLM architecture. Left: In a MHA module, the Key (K) and Value (V) sequences can be provided either internally (self-MHA) or externally (cross-MHA). Right: Cross-MHA is unavailable in the transformer block of decoder-only LLMs.}
	\vspace{-0.1in}
	\label{fig:attention}
\end{wrapfigure}

\subsection{Entry Points on Transformer Blocks}
Current LLMs [\citenum{zhang2022opt,scao2022bloom,chung2022scaling,touvron2023llama}] are stacked by transformer blocks [\citenum{vaswani2017attention}], each of which contains Multi-Head Attention (MHA) modules, LayerNorms, and a Feed-Forward Network (FFN). As shown in Figure \ref{fig:attention} - Left, MHA receives $H$ segments of Key (K), Value (V), and Query (Q) token sequences and performs attention through $H$ heads in parallel. The scaled dot-product attention mimics database search by weighted-summing the Values using their Keys' similarity to the Query. For self-MHA, K, V, and Q come internally from the previous LLM block's output. 

An intuitive solution to connecting an encoder to the LLM is to use the K-V pairs in cross-MHA. However, cross-MHA is only available in encoder-decoder LLMs [\citenum{chung2022scaling}], as shown in Figure \ref{fig:attention} - Right. The decoder-only LLMs, as the dominant LLM architecture [\citenum{chowdhery2022palm}], do not contain pre-trained MHA modules for cross-attention purposes\footnote{Although it is possible to re-purpose the self-MHA for cross-MHA by using external K-V pairs, doing so requires expensive retraining and changing the LLM structure at runtime.}. Alternatively, existing prompt tuning methods prepend trainable tokens to the input text [\citenum{lester2021power}] or intermediate K-V pairs [\citenum{li2021prefix, liu2022p}], but they only apply to the text domain.

Inspired by these existing efforts, we further expand the scope of adaptation to multiple modalities, by projecting multimodal tokens via plug-in projectors and then inserting them as new K-V pairs into MHA modules of LLM blocks.

\subsection{Efficient Insertion of External Modalities}
After having connected encoders to LLM's transformer blocks, there are multiple choices of inserting multimodal information from encoders to LLM. Insertion into the LLM's latent space enables more efficient cross-modal interaction [\citenum{nagrani2021attention, shukor2023ep}] than only into the input layer, due to more opportunities to interact with different levels of feature representations. Early work adopts FiLM-like weighting schemes [\citenum{de2017modulating, perez2018film, hu2018explainable, brohan2022rt}] to condition one modality on others in intermediate layers, but only hard-coded two modalities and require domain-specific structure designs from scratch. 
Later schemes used cross-MHA mechanisms for cross-modal interaction [\citenum{tan2019lxmert, tang2022tvlt, li2023trocr, xu2023bridgetower}], but cannot be applied on decoder-only LLMs. To avoid changing the pre-trained LLM structures, recent work [\citenum{shukor2023ep}] directly concatenates the projected multimodal tokens with output sequences from multiple LLM blocks. However, LLM blocks for such concatenation are arbitrarily selected and not optimized for training accuracy or speed. Extending the output sequence length for one LLM block will also affect the token sequence length in other blocks and reduce accuracy. 

These limitations of existing schemes motivate us to design a more flexible and generic method to insert multimodal information into pre-trained LLMs, and our basic idea is to allow adaptive weighting of multimodal tokens being inserted into the K-V set in  LLM blocks. In this way, we can establish optimal connections between encoders and LLM at runtime, for efficient cross-modal interaction.


%

\begin{wrapfigure}{R}{3in}
	\centering
	\includegraphics[width=3in]{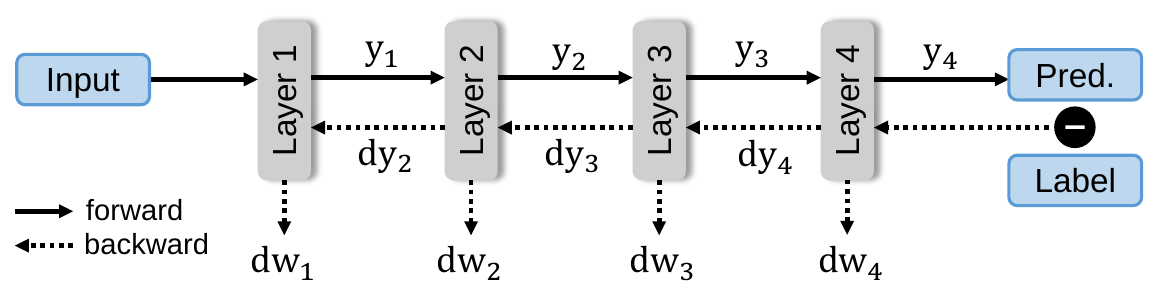}
	\vspace{-0.15in}
	\caption{Backpropagation of a 4-layer dense NN}
	\vspace{-0.1in}
	\label{fig:backprop}
\end{wrapfigure}

\subsection{FLOPs Model for Modality Adaptation}
\label{subsec:flops_model}
Our approach is inspired by the backpropagation model for modality adaptation. Typically, the FLOPs of backpropagation can be decomposed into two parts using the chain rule. As shown in Figure \ref{fig:backprop}, when training a 4-layer dense neural network, each layer computes $i)$ the activation gradient $\mathrm{dy_i}$ and passes it backward, and $ii)$ computes the weight update $\mathrm{dw_{i}}$ using $\mathrm{dy_{i+1}}$ from the upstream layer. Freezing some layers can eliminate the FLOPs of computing these layers' weight updates, but activation gradients will still be computed. For example, when freezing layer 2 to 4, the activation gradients from $\mathrm{dy_{2}}$ to $\mathrm{dy_{4}}$ still need to be computed for layer 1 to compute its weight update. Due to the generality of the chain rule, this mechanism applies to any other types of models, including transformer blocks in LLMs.

Existing work inserts new modalities into LLM's input layer through trainable projectors, and the inserted projectors can be considered as the model's first layer. Even when all the LLM layers are frozen, the projector still needs activation gradients to be passed through the entire LLM. Instead, our design connects input modalities into the last layers of LLM, to minimize the training cost at runtime. 

\begin{figure*}[ht]
	\centering
	\hspace{0.05in}
	\vspace{-0.1in}
	\includegraphics[width=5.5in]{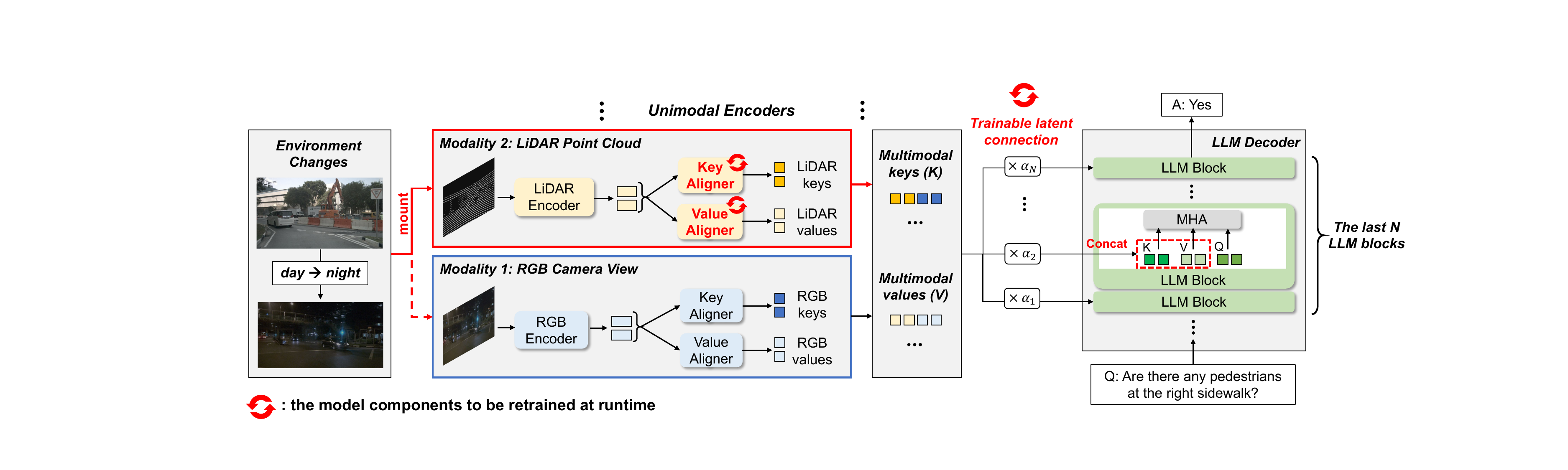}
	\vspace{-0.1in}
	\caption{An example of mPnP-LLM's modality adaptation for the multimodal QA task between two input modalities, namely RGB camera view and LiDAR point cloud. The text tokenizer, detokenizer, input and output embedding layers of the LLM are omitted for simplicity.}
	\label{fig:overview}
\end{figure*}

\section{mPnP-LLM Design}
We first discuss possible design choices of mPnP-LLM, and then present technical details of main modules in our design. Our design is generic and applicable to different types of decoder-only LLMs and generative tasks. In this paper, for simplicity, we use the multimodal question-answering (QA) task in autonomous driving [\citenum{qian2023nuscenes}] as an illustrative example of embodied AI applications, as shown in Figure \ref{fig:overview}.

\noindent \textbf{Choices of unimodal encoders.} Many pre-trained transformers, such as ViT [\citenum{dosovitskiy2020image}], AST [\citenum{gong2021ast}], Autoformer [\citenum{wu2021autoformer}], and Point-BERT [\citenum{yu2022point}], can be used as unimodal encoders to extract modality-specific features into token representations. With the token representations extracted from transformers, we only need to reshape the embedding dimension when designing projectors for inserting new modalities into the LLM. Other legacy encoder architectures based on MLP, CNN and LSTM can also be adopted by mPnP-LLM, but require additional efforts on reshaping their feature vectors or feature maps to token representations for LLM.


\noindent \textbf{Feature representation.} Most transformer-based encoders extract features into the [CLS] token sequences, and we further follow recent studies [\citenum{shukor2023ep}] to extract [CLS] tokens from multiple intermediate layers of encoders, to form a multi-level feature representation for more opportunities to align with the text features. We are also aware that not all the transformer-based encoders contain usable [CLS] tokens, which must be trained on relevant classification tasks. For other transformers, such as the ones pre-trained using Masked Auto-Encoder (MAE) [\citenum{he2022masked, hess2023masked}] or other non-classification tasks [\citenum{li2022exploring,ando2023rangevit}], we will use the averaged-pooled tokens for feature representation.

\vspace{0.05in}
\noindent Based on these design choices, mPnP-LLM adapts new input modalities and connects the corresponding unimodal encoders to LLM blocks via the following modules:
\begin{itemize}
	\item \textbf{Key \& Value Aligners}: We first use the trainable aligners to project multimodal tokens from the unimodal encoders into K-V pairs, which should be aligned with the text K-V pairs in the LLM blocks. Existing linear projections lack expressivity to ensure accurate alignment, and we address this challenge by introducing non-linearity in the aligner's projection with negligible computing cost.
	\item \textbf{Trainable Latent Connection}: Instead of injecting the aligned K-V pairs to every LLM block, we connect aligners to a flexible set of last blocks in LLM to save the runtime training cost, and use a lightweight post-weighting module to optimize the amount of multimodal information that flows to each LLM block from the same aligner. 
\end{itemize}
In practice, to ensure that the LLM can recognize newly adapted modalities, we first offload part of adaptation efforts, such as initial connections and adaptation to prompt structures, to offline training. Then at runtime, modality adaptation is performed by retraining the corresponding aligners and latent connections using the training data in new modalities, with the LLM and all encoders frozen.



\subsection{Key \& Value Aligners}
For modality adaptation, the encoder is connected to key and value aligners. For example in Figure \ref{fig:overview}, when switching the input modality, the new aligner is trained to project the unimodal tokens extracted from the LiDAR encoder into key-value pairs, which are then aligned with the key-value representations in LLM's latent space. Each aligner is a point-wise FFN with two linear transformations and a GELU activation function [\citenum{hendrycks2016gaussian}] in between:
\begin{equation}
\mathrm{Aligner}(x) = \mathrm{GELU}(xW_1 + b_1)W_2 + b_2,
\end{equation}
where $W_1, W_2, b_1, b_2$ are trainable parameters. This architecture is similar to the FFN in most transformer blocks, but our purpose is to align the token representation and match its shape to the LLM. Compared to a single linear transformation in the existing work [\citenum{zhu2023minigpt, shukor2023ep}], our design has stronger representation power because of the nonlinearity of activation functions, while adding negligible computing overhead compared to the high complexity of LLMs.


\subsection{Trainable Latent Connection}


The aligner's output of unimodal key and value tokens are concatenated into multimodal keys and values:
\begin{align*}
\mathbf{K}_{\mathrm{mm}} &= [K_1; K_2; ..., K_M] + \mathbf{T^{\mathrm{pos}}_{k}} \\
\mathbf{V}_{\mathrm{mm}} &= [V_1; V_2; ..., V_M] + \mathbf{T^{\mathrm{pos}}_{v}}
\end{align*}
where $\{K_{j}\}, \{V_{j}\}$ are unimodal key-value pairs and $M$ is the number of new modalities. Adding the learnable positional embedding $\mathbf{T^{\mathrm{pos}}_{k,v}}$ is optional, depending on the task complexity. For example in multimodal QA tasks, to answer questions like ``Is there any pedestrian on my left side?'', such positional embedding is necessary when the input data modality contains multiple parts (e.g., the RGB cameras on an autonomous vehicle may provide multiple views around the vehicle), for the LLM to recognize which token corresponds to which data part.

As shown in Figure \ref{fig:overview}, to inject multimodal information into $N$ LLM blocks, $\mathbf{K}_{\mathrm{mm}}$ and $\mathbf{V}_{\mathrm{mm}}$ are first duplicated $N$ times and then each duplication is multiplied by a trainable weight to tune its intensity. Specifically, such weighting procedure can be represented as:  
\begin{align*}
\mathbf{K'}_{\mathrm{mm}} &= [\alpha_1 \mathbf{K}_{\mathrm{mm}}, \alpha_2 \mathbf{K}_{\mathrm{mm}}, ..., \alpha_N \mathbf{K}_{\mathrm{mm}}] \\
\mathbf{V'}_{\mathrm{mm}} &= [\alpha_1 \mathbf{V}_{\mathrm{mm}}, \alpha_2 \mathbf{V}_{\mathrm{mm}}, ..., \alpha_N \mathbf{V}_{\mathrm{mm}}]
\end{align*}
where each $\alpha_j = \mathrm{sigmoid}(w_j / T)$ is parameterized by a trainable weight $w_j$ and a sigmoid function to ensure that its value is between 0 and 1, and we use the temperature $T$ as a hyper-parameter to control the sharpness of $\alpha_j$. Smaller $T$ makes the learned $\alpha_j$ more binary. The existing work [\citenum{jang2016categorical, gao2021network}] usually prefers to learn an exact binary representation of $\alpha_j$ using straight-through estimator [\citenum{bengio2013estimating}], but may reduce the representation power and cause unstable training. Instead, we aim to enable continuous tuning of the intensity of the injected multimodal information, so that each LLM block can flexibly retrieve different amounts of multimodal information by its need. 

The weighted multimodal key-value pairs are then concatenated with the projected text key-value pairs in the corresponding LLM block. For LLM block $j$, the key-value pairs for MHA are extended as:
\begin{align*}
\mathbf{K}_{\mathrm{ext}} &= [\alpha_j \mathbf{K}_{\mathrm{mm}}; \mathbf{K}_{\mathrm{txt}}] \\
\mathbf{V}_{\mathrm{ext}} &= [\alpha_j \mathbf{V}_{\mathrm{mm}}; \mathbf{V}_{\mathrm{txt}}].
\end{align*}

When $\alpha_j$ = 1, the original information in $\mathbf{K}_{\mathrm{mm}}$ and $\mathbf{V}_{\mathrm{mm}}$ are fed without any decay to the LLM block. Otherwise, $\alpha_j = 0$ degrades the impact of such information to zero, as no external tokens are injected. This is because by looking at the MHA's attention calculation:
\begin{equation}
\mathrm{Out} = \mathrm{softmax}(\frac{\mathbf{Q}[\alpha_j \mathbf{K}_{\mathrm{mm}}; \mathbf{K}_{\mathrm{txt}}]^{\top}}{\sqrt{d}})[\alpha_j \mathbf{V}_{\mathrm{mm}}; \mathbf{V}_{\mathrm{txt}}].
\end{equation}

As long as $\alpha_j \mathbf{V}_{\mathrm{mm}}$ is zero, the involvement of $\mathbf{K}_{\mathrm{mm}}$ and $\mathbf{V}_{\mathrm{mm}}$ will be zeroed in MHA calculation. This property is important because it allows us to eliminate the impact of some modality's information if it impairs the LLM's inference accuracy. Weighting on both keys and values also avoids complex designs of attention masks.

Note that implementing such concatenation does not require any modification of the existing LLM's computing graph. Most popular transformer frameworks (e.g., HuggingFace [\citenum{wolf2019huggingface}]) have already implemented interfaces to pass $\alpha_j \mathbf{K}_{\mathrm{mm}}$ and $\alpha_j \mathbf{V}_{\mathrm{mm}}$ to perform concatenation internally. The user only needs to extend the attention mask, so that the LLM can attend to the inserted keys and values.

\subsection{Deciding the Number of Connections}
As illustrated in Figure \ref{fig:overview}, such latent connection injects the multimodal information to the last $N$ LLM blocks, and the training cost hence depends on the value of $N$. Specifically, the backpropagation FLOPs can be calculated as:
\begin{equation}
T_{\mathrm{backprop}}(N) = T_{\mathrm{dw}}^{\mathrm{Aligners}} + T_{\mathrm{dy}}^{\mathrm{Emb}} + \frac{N}{L} T_{\mathrm{dy}}^{\mathrm{LLM}},
\end{equation}
where $T_{\mathrm{dw}}^{\mathrm{Aligners}}$ is the FLOPs to compute weight updates of aligners, and $T_{\mathrm{dy}}^{\mathrm{Emb}}$ is the FLOPs to pass activation gradients through the LLM's output embedding layer, and $T_{\mathrm{dy}}^{\mathrm{LLM}}$ is the FLOPs to pass activation gradients through all the LLM blocks. In practice, due to much higher complexity of the LLM blocks, we have $T_{\mathrm{dw}}^{\mathrm{Aligners}} + T_{\mathrm{dy}}^{\mathrm{Emb}} \ll T_{\mathrm{dy}}^{\mathrm{LLM}}$. The ratio of training cost with respect to connecting to all $L$ LLM blocks can be approximated as:
\begin{equation}
r = \frac{T_{\mathrm{backprop}}(L)}{T_{\mathrm{backprop}}(N)} \approx \frac{L}{N}.
\end{equation}

In this way, the user can flexibly control the training cost at runtime, by tuning $N$ based on different application requirements and system resource conditions.

\begin{algorithm}
	\caption{Modality Adaptation in mPnP-LLM}\label{alg:modaltiy_adaptation}
	\begin{algorithmic}
		\Require:	A set of pre-trained encoders $\mathbf{E} = \{E_1, E_2, ...\}$ and the corresponding aligners $\mathbf{A} = \{A_1, A_2, ...\}$ stored on local external storage; a pre-trained LLM loaded in memory; trainable latent connections $\alpha_{1,...,N}$
		\\\hrulefill
		
		\State {\color{teal}\text{/* Offline preparation */}}
		\State $\mathbf{E}_0, \mathbf{A}_0 \gets \texttt{Select}(\mathbf{E}, \mathbf{A})$ \Comment{Load initial encoders}
		\State $\mathbf{LLM} \gets \texttt{Reconnect}(\mathbf{E}_0, \mathbf{A}_0)$ \Comment{Connect to LLM}
		\State $\texttt{Train}(\mathrm{LLM}_{k,v}, \mathbf{A}_0, \alpha_{1,...,N})$ \Comment{Offline training}
		
		\State {\color{teal}\text{/* Runtime modality adaptation */}}
		\While{$t < T_{\mathrm{end}}$}
		\State $\mathbf{E}_t, \mathbf{A}_t \gets \texttt{Select}(\mathbf{E}, \mathbf{A})$ \Comment{Reload encoders}
		\State $\mathbf{LLM} \gets \texttt{Reconnect}(\mathbf{E}_t, \mathbf{A}_t)$ \Comment{Reconnect}
		\State $\texttt{Train}(\mathrm{LLM}_{k_{1,...,N},v_{1,...,N}}, \mathbf{A}_t, \alpha_{1,...,N})$ \Comment{Adapt}
		\EndWhile
	\end{algorithmic}
\end{algorithm}

\subsection{Practical Workflow for Modality Adaptation}
With aligners and trainable latent connections described above, modality adaptation can be performed with proper workflows. When being assigned a task, such as multimodal QA, the embodied AI device should have multiple unimodal encoders for different input modalities and their pre-trained aligners being stored on the device's local external storage. We assume that the user has prior knowledge to select the relevant modalities that are initially connected to the LLM, and that the embodied AI system can detect the environmental changes and automatically select new modalities for adaptation. In the example shown in Figure \ref{fig:overview}, we assume that the embodied AI system starts with the modality of RGB camera view by default and will automatically adapt to the modality of LiDAR point cloud in nighttime.

As shown in Algorithm \ref{alg:modaltiy_adaptation}, the initial connection is pre-trained offline with a base dataset that is relevant to the task. For example, to perform domain-specific QA tasks, the model can be pre-trained on generic but small-scale datasets (e.g., OKVQA [\citenum{marino2019ok}]) to gain basic QA capabilities. In particular, all the key and value projectors in the LLM are also trained to quickly adapt to input and output text prompt structures, such as ``Given the question: {Q}, the answer is:''.

After offline preparation, the system can be used for runtime modality adaptation, by retraining the aligners and latent connections at runtime with the newly inserted encoders. Besides, it is also still necessary to update the K, V projectors in the connected LLM blocks, so that the LLM can distinguish newly inserted tokens from the existing ones. To minimize the training cost in this part, we apply LoRA [\citenum{hu2021lora}] because it introduces negligible FLOPs to compute weight updates. Note that we only apply LoRA to the connected blocks, which is different from the existing work [\citenum{doveh2023teaching, gong2023multimodal}] that applies LoRA to all the LLM blocks for excessive representation power.

\begin{table*}
	\centering
	{\fontsize{8}{9}\selectfont
		\begin{tabular}{lrrrrrr}
			\toprule
			\multirow{2}{*}{\textbf{Method}} & \multicolumn{4}{c}{\textbf{Accuracy (\%) w.r.t Scene \& Modality}} & \multicolumn{2}{c}{\textbf{Cost w.r.t Night (C $\rightarrow$ L)}} \\
			\cmidrule(lllr){2-5} \cmidrule(llr){6-7}
			& \textbf{Day (C)} & \textbf{Night (C)} & \textbf{Night (C $\rightarrow$ C + L)} & \textbf{Night (C $\rightarrow$ L)} & \textbf{PFLOPs} & \textbf{Memory (GB)} \\
			\midrule
			Full LLM & 32.5 & 30.2 & 3.5 & 3.5 & 1.58 & 29.9 \\
			PromptFuse & 33.3 & 26.9 & 24.9 & 39.2 & 1.09 & 26.0 \\
			eP-ALM & 34.7 & 24.6 & 36.0 & 44.7 & 1.11 & 27.6 \\
			mPnP-LLM ($N=4$) & 25.2 & 21.1 & 22.3 & 24.9 & 0.68 & 23.2 \\
			mPnP-LLM ($N=7$) & 40.1 & 34.1 & 25.6 & 26.3 & 0.74 & 23.9 \\
			mPnP-LLM ($N=10$) & 49.1 & 41.1 & 27.0 & 41.9 & 0.81 & 24.5 \\
			mPnP-LLM ($N=13$) & 49.2 & 40.1 & 38.5 & 46.4 & 0.87 & 23.1 \\
			mPnP-LLM ($N=16$) & 50.3 & 43.4 & 41.7 & 46.9 & 0.93 & 25.9 \\
			mPnP-LLM ($N=19$) & 50.7 & 41.0 & 44.0 & 46.1 & 0.99 & 23.9 \\
			mPnP-LLM ($N=22$) & 47.8 & 39.2 & 41.7 & 47.5 & 1.05 & 26.0 \\
			\bottomrule
	\end{tabular}}
	\vspace{-0.05in}
	\caption{Performance of mPnP-LLM vs. baseline schemes w.r.t scenes (Day/Night) and modalities (C: 6 RGB camera views, L: LiDAR point cloud) using OPT-1.3B by connecting with different numbers of LLM blocks ($N$). The OPT-1.3B model has 24 LLM blocks in total.}
	\label{tab:connection_count}
\end{table*}

\section{Experiments}
\noindent \textbf{Dataset preparation and evaluation setup.} We use the nuScenes-QA dataset for multimodal visual QA in autonomous driving [\citenum{caesar2020nuscenes}]. The original nuScenes-QA dataset is too large ($\sim$460k QA pairs) for runtime training. Instead, we build a dataset named nuScenes-QA-mini on top of nuScenes-QA benchmark [\citenum{qian2023nuscenes}], as the overlap part between the nuScenes-QA dataset and the mini-split of the original nuScenes database. 

We consider a challenging evaluation scenario of runtime modality adaptation: the model should switch from RGB camera view in daytime to LiDAR point cloud in nighttime. We divide the nuScenes-QA-mini dataset into day (4,458 samples) and night (1,138 samples) splits based on nuScenes' metadata. These two splits are further divided by half for training and testing, respectively. More details about datasets, as well as encoders and LLMs used in evaluations, can be found in Section A of the Appendix.

\noindent \textbf{Baseline schemes.} For runtime modality adaptation, existing modular and parameter-efficient multimodal learning schemes are competitive with mPnP-LLM. They adopt trainable projectors and adapters (e.g., prompt tuning [\citenum{lester2021power}]) for plug-and-play encoders. We compare mPnP-LLM with the following three baselines\footnote{Note that, we did not include those methods that hard-code the model structure for adapting to a particular downstream task [\citenum{perez2018film,wang2022git}], because their designs lack flexibility for modality adaptation and can only change modality by redesigning and training a new model from scratch.}:
\begin{itemize}
	\item \textbf{Full LLM} [\citenum{brohan2022rt, driess2023palm}]: It connects multimodal encoders with the LLM input layer, and trains the projector and fine-tunes the entire LLM for modality adaptation. It is the most expensive and data-hungry method, but can potentially lead to the highest task accuracy.
	\item \textbf{PromptFuse} [\citenum{liang2022modular}]: Similarly, it connects the encoders with the input layer of the LLM. To reduce the training cost of modality adaptation, instead of fine-tuning all the parameters, it adopts prompt tuning to adapt the LLM.
	\item \textbf{eP-ALM} [\citenum{shukor2023ep}]: It applies hard-coded connections between encoders' [CLS] tokens and intermediate LLM blocks, and uses prompt tuning to adapt the LLM. Since the LiDAR encoder does not provide well-trained [CLS] tokens, we feed its average-pooled output tokens to LLM, being consistent with our implementation in mPnP-LLM.
\end{itemize}
\vspace{0.05in}

We implemented mPnP-LLM in PyTorch, and conducted our experiments on 1) a Dell workstation with a Nvidia RTX A6000 48GB GPU and 2) an edge device of Nvidia Jetson AGX Orin with a Nvidia Ampere GPU of 64 tensor cores and 64GB shared memory. We use the \emph{exact match ratio} as the accuracy metric to indicate how many generated answers exactly match the ground truth. More implementation details can be found in Section B of the Appendix.

\subsection{Modality Adaptation Cost \& Accuracy}
We first compare mPnP-LLM with other baseline schemes on the OPT-1.3B model. In the evaluation scenario described above, the model is first offline trained on the day-train split to connect the RGB modality to LLM, and tested on the day-test split where the test accuracy corresponds to Day (C) in Table \ref{tab:connection_count}. The trained model is then tested on the night-test split to show the failure of RGB modality in nighttime conditions, where the test accuracy corresponds to Night (C). Then, two types of modality adaptations are evaluated on the night-test split: 1) Night (C $\rightarrow$ C + L) mounts the LiDAR modality and still keeps the RGB modality; 2) Night (C $\rightarrow$ L) mounts the LiDAR modality but detaches the RGB modality.

Results in Table \ref{tab:connection_count} show that in offline training - Day (C), mPnP-LLM achieves 5\%-15\% higher test accuracy than baseline schemes. When being zero-shot evaluated on Night (C), the accuracy drops significantly and demonstrates the need of modality adaptation. After having adapted to the LiDAR modality, mPnP-LLM ($N=13,...,22$) can resume the task accuracy by up to 8\%, but uses 20\%-45\% less training FLOPs and 25\% less GPU memory compared to baseline schemes. Including both RGB and LiDAR modalities in nighttime (C $\rightarrow$ C + L) surprisingly impairs the accuracy, possibly because keeping the uninformative RGB modality confuses LLM inference.

Note that the task accuracy with different input modalities could be inherently different, depending on the knowledge being provided. For example, RGB images contain fine-grained object information (e.g., color and texture). LiDAR point clouds can be relatively sparse and only provide information about distances and 3D object structures. Hence, the accuracy with the LiDAR modality at nighttime is lower than that with the RGB modality at daytime.

\begin{figure}
	\centering
	\vspace{-0.2in}
	\hspace{-0.25in}
	\subfigure[OPT-1.3B]{
		\centering
		\includegraphics[width=2.2in]{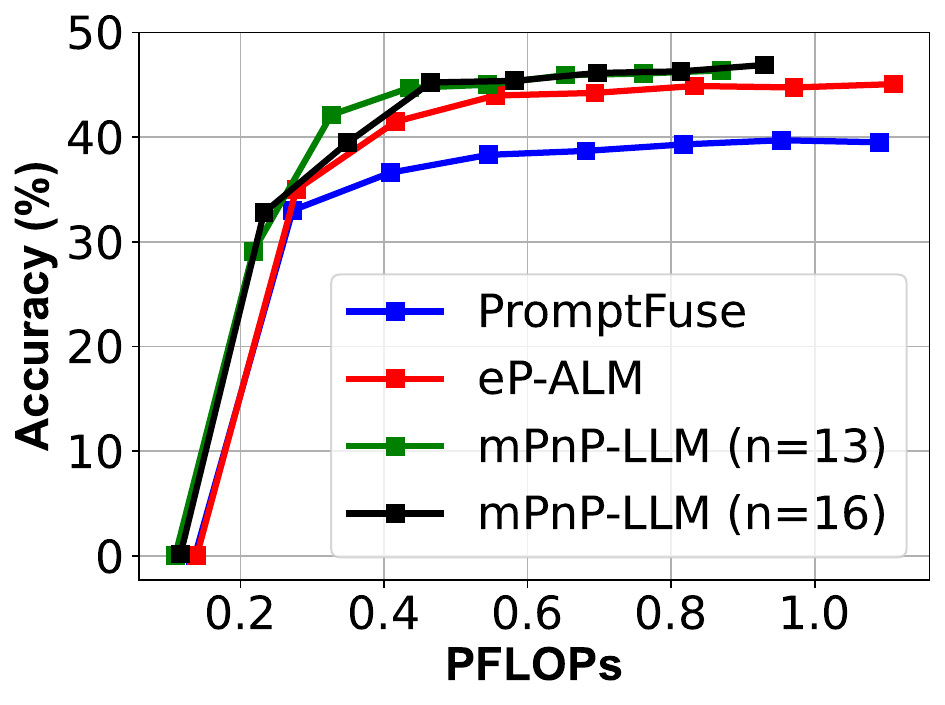}
		\label{fig:acc_flops_opt1b3}
	}
		\hspace{0.25in}
	\subfigure[OPT-2.7B]{
		\centering
		\includegraphics[width=2.2in]{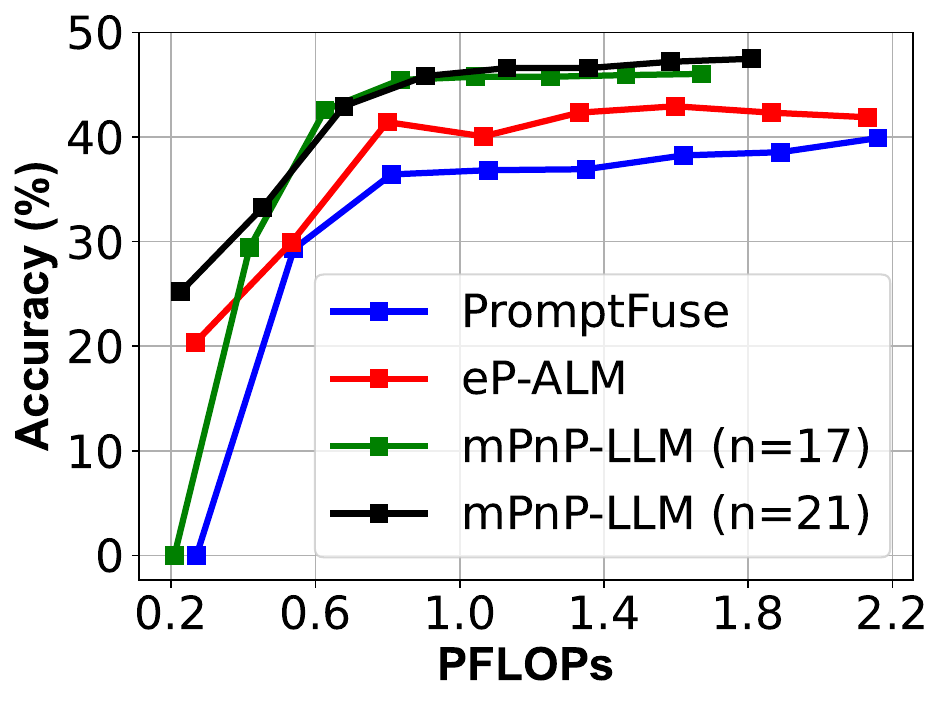}
		\label{fig:acc_flops_opt2b7}		
	}
	\vspace{-0.1in}
	\caption{Accuracy-compute tradeoff w.r.t Night (C $\rightarrow$ L)}
	\vspace{-0.2in}
	\label{fig:acc_flops}
\end{figure}

Figure \ref{fig:acc_flops} further shows the accuracy-compute efficiency of mPnP-LLM, as how the task accuracy evolves with respect to the amount of FLOPs being incurred, when adapting C $\rightarrow$ C + L. mPnP-LLM achieves better accuracy-compute tradeoff compared to baselines. Specifically, it reduces the FLOPs by 3.6$\times$ to achieve 40\% accuracy compared to PromptFuse, and reduces the FLOPs by 1.7$\times$ to achieve 45\% accuracy compared to eP-ALM. Such performance margin is further enlarged when using larger LLMs. On OPT-2.7B, to achieve the same final accuracy as eP-ALM, mPnP-LLM reduces the training FLOPs by 3.7$\times$. This good accuracy-compute tradeoff stems from both the improved convergence and less backpropagation cost.

\begin{table}[ht]
	\centering
	{\fontsize{8}{9}\selectfont
		\begin{tabular}{lrr}
			\toprule
			\textbf{Ablated module} & \textbf{Accuracy ($N=13$)} & \textbf{Accuracy ($N=17$)} \\
			\midrule
			None & 42.2 & 46.0 \\
			Offline train K,V proj. & 37.8 & 42.9 \\
			Aligner: MLP$\rightarrow$Linear & 37.3 & 43.8 \\
			Conn.: trained $\rightarrow$ fixed  & 39.5 & 44.7 \\
			LoRA on K,V proj. & 33.8 & 35.4 \\
			\bottomrule
	\end{tabular}}
	\vspace{0.1in}
	\caption{Ablation study of mPnP-LLM with OPT-2.7B as the LLM}
	\label{tab:ablation}
\end{table}

\subsection{Ablation Study}
An ablation study in Table \ref{tab:ablation} showed that each component in mPnP-LLM contributes at least 3\% to the task accuracy, compared to accuracy without ablation. Disabling the LoRA adapter causes the highest accuracy drop, because modality adaptation requires updating the K, V projectors to correctly distinguish newly inserted tokens from existing tokens. Similar phenomena are observed in the existing work [\citenum{shukor2023ep}] when adopting LLM adapters. The accuracy drop due to ablation is lower when connecting to more LLM blocks, because higher representation power is present when the multimodal tokens go through more LLM blocks.


\begin{table*}[ht]
	\centering
	{\fontsize{8}{9}\selectfont
		\begin{tabular}{lrrrrrrr}
			\toprule
			\multirow{2}{*}{\makecell{\textbf{LLM} \\ \textbf{\& Method}}} & \multicolumn{4}{c}{\textbf{Accuracy (\%) w.r.t Scene \& Modality}} & \multicolumn{2}{c}{\textbf{Cost w.r.t Night (C $\rightarrow$ L)}} \\
			\cmidrule(lllr){2-5} \cmidrule(llr){6-7}
			& \textbf{Day (C)} & \textbf{Night (C)} & \textbf{Night (C $\rightarrow$ C + L)} & \textbf{Night (C $\rightarrow$ L)} & \textbf{PFLOPs} & \textbf{Memory (GB)} \\
			\midrule
			\rowcolor{gray!25}
			\vspace{0.05in}
			\textbf{OPT-350M} \\
			PromptFuse & 27.7 & 18.8 & 0.0 & 34.9 & 0.32 & 21.1 \\
			eP-ALM & 25.3 & 13.5 & 24.1 & 36.9 & 0.33 & 20.0 \\
			mPnP-LLM ($N=10$) & 44.0 & 37.5 & 26.1 & 38.5 & 0.25 & 19.8 \\
			mPnP-LLM ($N=13$) & 45.4 & 36.6 & 31.6 & 37.7 & 0.26 & 20.0 \\
			mPnP-LLM ($N=16$) & 45.9 & 37.8 & 31.7 & 40.5 & 0.28 & 21.0 \\
			\midrule
			\rowcolor{gray!25}
			\vspace{0.05in}
			\textbf{OPT-1.3B} \\
			PromptFuse & 33.3 & 26.9 & 24.9 & 39.2 & 1.09 & 26.0 \\
			eP-ALM & 34.7 & 24.6 & 36.0 & 44.7 & 1.11 & 27.6 \\
			mPnP-LLM ($N=10$) & 49.1 & 41.1 & 27.0 & 41.9 & 0.81 & 24.5 \\
			mPnP-LLM ($N=13$) & 49.2 & 40.1 & 38.5 & 46.4 & 0.87 & 23.1 \\
			mPnP-LLM ($N=16$) & 50.3 & 43.4 & 41.7 & 46.9 & 0.93 & 25.9 \\
			\midrule
			\rowcolor{gray!25}
			\vspace{0.05in}
			\textbf{OPT-2.7B} \\
			PromptFuse & 35.7 & 26.1 & 28.8 & 39.8 & 2.16 & 36.4 \\
			eP-ALM & 37.3 & 24.7 & 24.9 & 41.9 & 2.13 & 36.4 \\
			mPnP-LLM ($N=13$) & 50.2 & 36.9 & 31.3 & 42.2 & 1.56 & 28.4 \\
			mPnP-LLM ($N=17$) & 51.2 & 37.2 & 27.6 & 46.0 & 1.67 & 30.1 \\
			mPnP-LLM ($N=21$) & 52.3 & 42.2 & 44.3 & 46.4 & 1.81 & 28.9 \\
			\bottomrule
	\end{tabular}}
	\vspace{-0.05in}
	\caption{Performance of mPnP-LLM vs. baseline schemes using OPT-350M, 1.3B and 2.7B models, by connecting with different numbers of LLM blocks ($N$).}
		\vspace{-0.1in}
		\label{tab:llm_size}
	\end{table*}

\subsection{Impact of LLM Size}
We compare mPnP-LLM with baseline schemes using OPT-350M, OPT-1.3B and OPT-2.7B models. We keep the ratio of connected LLM blocks, which decides the model's representation power in newly added modalities, to be consistent across different models. For example, we set $N=10\sim16$ for OPT-350M with 24 blocks, and set $N=13\sim21$ for OPT-2.7B with 32 blocks, leading to a connection ratio between 41\% 66\% for both.

As shown in Table \ref{tab:llm_size}, accuracy improvements of mPnP-LLM ($r=42\%$) are higher on OPT-350M than on bigger models. With OPT-2.7B, eP-ALM suffers a significant accuracy drop on Night (C $\rightarrow$ L), while mPnP-LLM can retain performance compared to the OPT-1.3B cases. We attribute this to mPnP-LLM's trainable connections that can adaptively reconnect for different LLM sizes. Essentially, with different values of $N$, the training costs of mPnP-LLM have been very close to the backpropagation costs of encoders and LLM blocks. Hence, training the aligners and latent connections in mPnP-LLM are very lightweight. 

The peak GPU memory usage in mPnP-LLM is mainly decided by the LiDAR encoder, since it directly takes high-volume point cloud tensors as input. LLM's memory consumption is relatively lower, due to its small context window (64 tokens). Since mPnP-LLM freezes the encoders in runtime training, its memory savings only stem from the smaller backpropagation cost and are hence limited to 20\% when the LLM size is small (350M and 1.3B). However, the memory savings increase with larger models and larger context windows. On OPT-2.7B, the memory savings can increase by another 10-15\%.

\begin{table}[ht]
	\centering
	{\fontsize{8}{9}\selectfont
		\begin{tabular}{lrc}
			\toprule
			{\makecell{\textbf{LLM} \\ \textbf{\& Method}}} & {\makecell{\textbf{Day (C)} \\ \textbf{Acc. (\%)}}} & {\makecell{\textbf{Night (C $\rightarrow$ L)} \\ \textbf{Acc. (\%) / PFLOPs / Mem. (GB)}}} \\
			\midrule
			\rowcolor{gray!25}
			\vspace{0.05in}
			\textbf{OPT-1.3B} \\
			Full LLM & 32.5 & 3.5 / 1.58 / 29.9 \\
			PromptFuse & 33.3 & 39.2 / 1.09 / 26.0 \\
			eP-ALM & 34.7 & 44.7 / 1.11 / 27.6 \\
			mPnP-LLM ($n=10$) & 49.1 & 41.9 / 0.81 / 24.5 \\
			mPnP-LLM ($n=13$) & 49.2 & 46.4 / 0.87 / 23.1 \\
			
			\midrule
			\rowcolor{gray!25}
			\vspace{0.05in}
			\textbf{BLOOMZ-1.1B} \\
			Full LLM & 34.0 & 0.0 / 1.16 / 32.2 \\
			PromptFuse & 35.4 & 26.6 / 0.90 / 25.7 \\
			eP-ALM & 27.1 & 26.0 / 0.91 / 28.7 \\
			mPnP-LLM ($n=10$) & 39.4 & 25.8 / 0.73 / 22.7 \\
			mPnP-LLM ($n=13$) & 44.0 & 27.0 / 0.76 / 25.8 \\
			\bottomrule
	\end{tabular}}
	\vspace{0.1in}
	\caption{Task accuracy \& training cost on different LLMs}
	\label{tab:llm_types}
	\vspace{-0.1in}
\end{table}

\subsection{Performance with Different Types of LLMs}
We also investigated the performance of mPnP-LLM with different types of LLMs, including both OPT and BLOOMZ. OPT is pre-trained on the generic text corpus and the pre-training of BLOOMZ focuses more on cross-lingual text generation, leading to different reasoning power. As shown in Table \ref{tab:llm_types}, the overall task accuracy on BLOOMZ is lower than OPT, because cross-lingual pre-training could reduce the LLM's performance on a specific language when the model complexity is low [\citenum{luo2023empirical}]. Nevertheless, mPnP-LLM still achieves higher accuracy than baseline schemes, and reduces the training FLOPs by 20\%-37\% and GPU memory consumption by up to 30\%.

\subsection{Impact of the Amount of Training Samples}
In our evaluation, the amount of runtime training samples is 659, which is very small compared to the size of existing QA datasets (e.g., 9k for OKVQA [\citenum{marino2019ok}] and 444k for VQA-v2 [\citenum{goyal2017making}]) to minimize the cost. As shown in Table \ref{tab:sample_count}, mPnP-LLM can achieve higher accuracy compared to the baseline schemes, when the amount of runtime training samples further reduces by up to 40\%. In particular, when using 527 training samples, mPnP-LLM can still achieve on-par accuracy with eP-ALM using 659 training samples, but further reduces the training FLOPs by 20\%.

\begin{table*}[ht]
	\centering
	{\fontsize{8}{9}\selectfont
		\begin{tabular}{lrrrrrrrr}
			\toprule
			\multirow{2}{*}{\textbf{Method}} & \multicolumn{2}{c}{\textbf{zero-shot}} & \multicolumn{2}{c}{\textbf{\# sample = 395 (60\%)}} & \\
			\cmidrule(ll){2-3} \cmidrule(ll){4-5} 
			& \textbf{Accuracy (\%)} & \textbf{PFLOPs} & \textbf{Accuracy (\%)} & \textbf{PFLOPs} \\
			\midrule
			PromptFuse & 26.9 & - & 37.3 & 0.65 \\
			eP-ALM & 24.6 & - & 41.5 & 0.67 \\
			mPnP-LLM ($N=13$) & 40.1 & - & 41.0 & 0.52 \\
			mPnP-LLM ($N=16$) & 43.4 & - & 43.9 & 0.56 \\	
			\cmidrule(ll){1-5}
			\multirow{2}{*}{\textbf{Method}} & \multicolumn{2}{c}{\textbf{\# sample = 527 (80\%)}} & \multicolumn{2}{c}{\textbf{\# sample = 659 (100\%)}} & \\
			\cmidrule(ll){2-3} \cmidrule(ll){4-5} 
			& \textbf{Accuracy (\%)} & \textbf{PFLOPs} & \textbf{Accuracy (\%)} & \textbf{PFLOPs} \\
			\midrule
			PromptFuse & 39.0 & 0.87 & 39.2 & 1.09 \\
			eP-ALM & 43.7 & 0.89 & 44.7 & 1.11 \\
			mPnP-LLM ($N=13$) & 44.0 & 0.70 & 46.4 & 0.87 \\
			mPnP-LLM ($N=16$) & 44.2 & 0.74 & 46.9 & 0.93 \\			
			\bottomrule
	\end{tabular}}
	\vspace{0.1in}
	\caption{The impact of the amount of training samples for runtime modality adaptation, on the OPT-1.3B model}
	\label{tab:sample_count}
\end{table*}

\subsection{Runtime Modality Adaptation on Edge Devices}
We further evaluated mPnP-LLM on a Nvidia Jetson AGX Orin [\citenum{orin}], which is widely used in embodied AI such as home intelligent robots [\citenum{hachibot}] and autonomous delivery [\citenum{meituan}]. As shown in Table \ref{tab:edge_device}, in its maximum power mode (MAXN), mPnP-LLM can iterate through the 2,636 training samples in our nuScenes-QA-mini dataset within 35 minutes, which corresponds to 4 epochs on the night-train split and resumes the task accuracy by 5\%. In comparison, the wall-clock time needed to complete the same training on workstation-level GPUs, such as NVidia RTX A6000, is 6 minutes.

\begin{table}[ht]
	\centering
	{\fontsize{8}{9}\selectfont
		\begin{tabular}{lrc}
			\toprule
			\textbf{Device setup} & \textbf{GPU Freq.} & \textbf{mPnP-LLM ($N=13$)} \\
			\midrule
			RTX A6000 300W & 1.9GHz & 9.40 samples/s\\
			AGX Orin 30W & 612MHz & 0.33 samples/s\\
			AGX Orin 50W & 816MHz & 0.92 samples/s\\
			AGX Orin MAXN & 1.3GHz & 1.41 samples/s\\
			\bottomrule
	\end{tabular}}
	\vspace{0.1in}
	\caption{Performance of runtime modality adaptation with OPT-1.3B on Nvidia Jetson AGX Orin}
	\vspace{-0.1in}
	\label{tab:edge_device}
\end{table}

Note that even without runtime modality adaptation, mPnP-LLM can retain a moderate level of zero-shot accuracy as shown in Table \ref{tab:sample_count}, which is at least 15\% higher than that of baselines and only 3\% lower than the task accuracy using 100\% training samples. Similarly, mPnP-LLM's task accuracy approximates to the peak accuracy even before it has iterated over all training samples. Such high accuracy ensures that the embodied AI system can smoothly adapt to new input modalities, and still use the RGB modality during such adaptation to retain the task performance.

\section{Conclusions}
In this paper, we present mPnP-LLM, a new technique that allows elastic runtime modality adaptation for multimodal LLMs in embodied AI. mPnP-LLM can achieve up to 3.7$\times$ FLOPs reduction while retaining on-par accuracy with the existing schemes. Under the same compute budget, mPnP-LLM improves the task accuracy by up to 4\% compared to the best existing scheme. More discussions about the expandability and generalizablity of mPnP-LLM are in Section C of the Appendix.

\setcitestyle{numbers}
\bibliographystyle{abbrvnat}
\bibliography{ref}

\section*{APPENDIX}
\appendix
\noindent In the appendix, we provide additional details and discussions that were included in the main texts of our paper. These include: 1) the details of the dataset that we used in evaluations; 2) the details about our evaluation setup and implementations of mPnP-LLM; and 3) discussions about technical design choices, expandability and generalizability of mPnP-LLM. The source codes of mPnP-LLM can be found at \url{https://github.com/pittisl/mPnP-LLM}.

\section{Details of Dataset used in Evaluations}



\begin{figure}[h]
	\centering
	\includegraphics[width=1.0\linewidth]{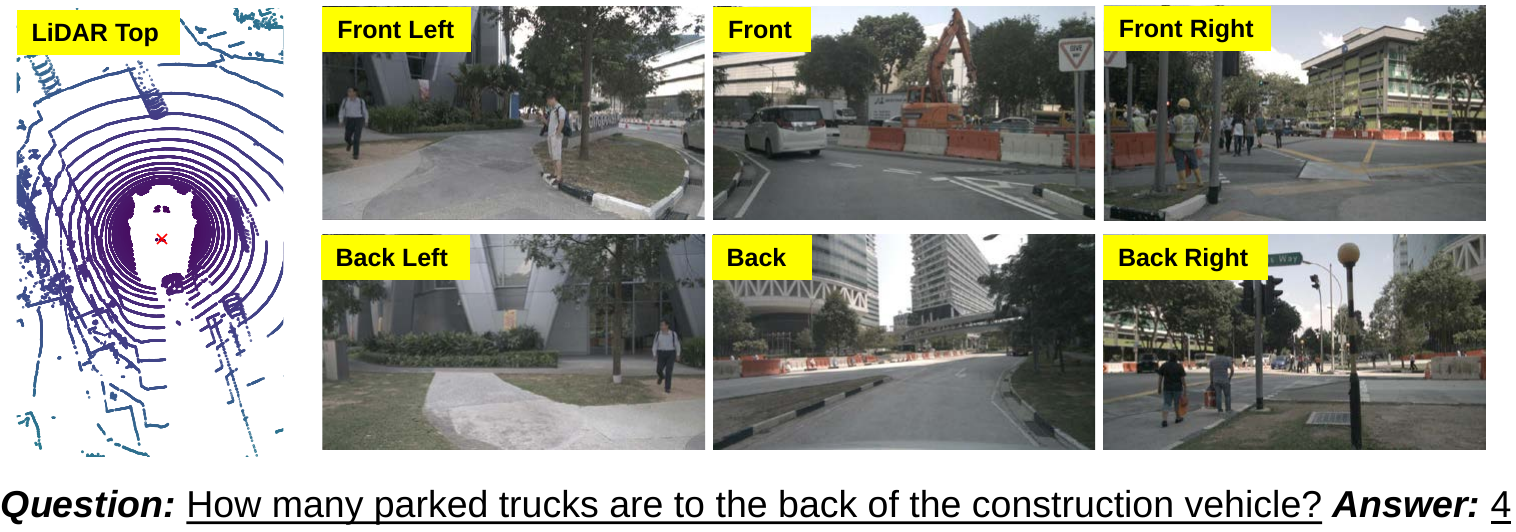}
	\vspace{-0.1in}
	\caption{A data sample from the nuScenes-QA dataset}
	\label{fig:nuqa_example}
\end{figure}

The dataset being used in our evaluations, namely nuScenes-QA-mini, is built on the nuScenes min-split [\citenum{caesar2020nuscenes}], where we obtain the QA pairs from the original nuScenes-QA dataset [\citenum{qian2023nuscenes}]. The data in the nuScenes-QA dataset is collected from driving scenes in cities of Boston and Singapore with diverse locations, time, and weather conditions. As shown in Figure \ref{fig:nuqa_example}, each data sample contains 6-view RGB camera captures, a 5D LiDAR point cloud, and a corresponding text QA pair. Each LiDAR point cloud includes 5 dimensions of data about distance, intensity, X, Y, and Z axes.

In this dataset, the questions are generally difficult, and may require multiple hops of reasoning over the RGB and LiDAR data. For example, to answer the sample question in Figure \ref{fig:nuqa_example}, the ML model needs to first identify in which direction the ``construction vehicle'' appears, and then counts the number of ``parked trucks'' in that direction. In our evaluations, we further cast the question-answering (QA) as an open-ended text generation task. This is more challenging than the evaluation setup in the original nuScenes-QA paper [\citenum{caesar2020nuscenes}], where an answer set is predefined and the QA task is a classification task over this predefined answer set.

\begin{figure}[h]
	\centering
	\includegraphics[width=1.0\linewidth]{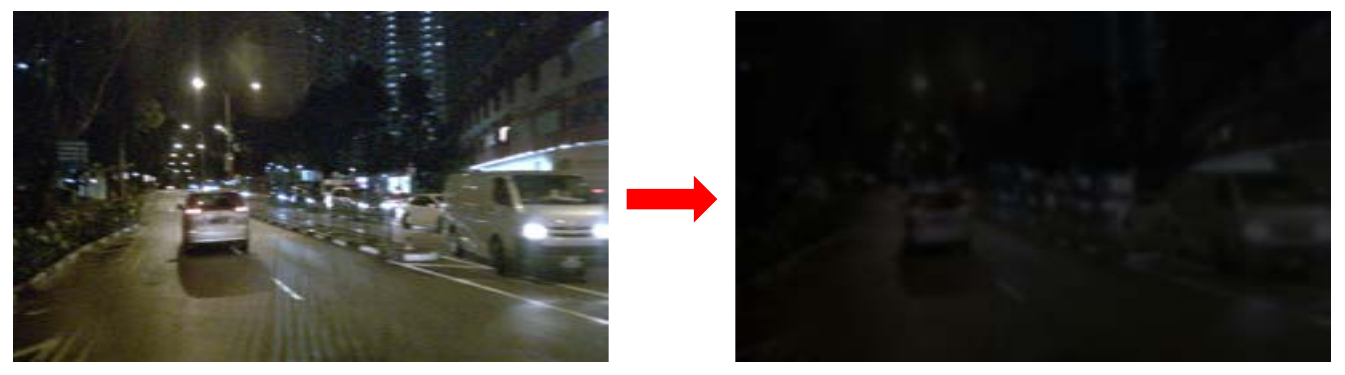}
	\vspace{-0.1in}
	\caption{An example of darkening and blurring the RGB view.}
	\label{fig:rgb_darken}
\end{figure}

In most RGB images in the original nuScenes-QA dataset, as shown in Figure \ref{fig:rgb_darken} - Left, the lighting conditions in night scenes are still abundant (e.g., with street lights). Hence, to create a challenging evaluation scenario in which modality adaptation from RGB camera view to liDAR point cloud is necessary in night scenes, we further reduce the brightness of RGB cameara captures in night scenes by 80\% and apply Gaussian blur with a radius of 7, as shown in Figure \ref{fig:rgb_darken} - Right. By applying such preprocessing to the RGB views in night-scene samples of our nuScenes-QA-mini dataset, we obtained the training and validation splits of night scenes, with 659 samples for each split. On the other hand, the RGB views in daytime scenes remain unchanged: the daytime split in the dataset contains 2,229 for training and 2,229 for validation, respectively. 

We built the dataset into Apache Arrow format for faster access at runtime, and provided the scripts of generating the nuScenes-QA-mini dataset from the original nuScenes-QA dataset in the source code repository. Please refer to the README.md file in the source code package for details.

\section{Details about Evaluation Setup and Implementation}

\noindent \textbf{Encoders and LLMs.} For the RGB camera view modality, we use a ViT-small [\citenum{dosovitskiy2020image}] encoder pre-trained on ImageNet-1k [\citenum{deng2009imagenet}] to extract RGB tokens. For each of the 6 RGB camera views in a data sample, we extract the [CLS] tokens from the last 4 transformer blocks of the encoder. For the LiDAR point cloud modality, we use a RangeViT [\citenum{ando2023rangevit}] encoder pre-trained on the nuScenes segmentation tasks to extract rich object-level information from the LiDAR point cloud. We average pool the RangeViT output into 4 tokens to represent features of the LiDAR modality. 

To perform QA tasks on multimodal data, we adopt pre-trained decoder-only LLMs, including OPT [\citenum{zhang2022opt}] and BLOOMZ [\citenum{muennighoff2022crosslingual}]. These LLMs offer different variants of parameter sizes (125M-176B), and in our evaluation, we select models with parameter sizes varying from 350M to 2.7B, to take different resource constraints of embodied AI devices into consideration.

\vspace{0.05in}
\noindent \textbf{Implementation.} We implemented mPnP-LLM in PyTorch, and conducted our experiments on 1) a Dell workstation with a Nvidia RTX A6000 48GB GPU and 2) a Nvidia AGX Orin edge device with a Nvidia Ampere GPU of 64 tensor cores and 64GB shared memory. The RGB images are resized to 224$\times$224 to match ViT-small's input requirement, and each LiDAR point cloud is projected to a 32$\times$2048$\times$3 tensor internally by RangeViT. Note that, since the amount of input data in the LiDAR point cloud modality is much larger than that in the RGB camera view modality, the LiDAR data consumes the majority of memory in runtime modality adaptation. The amount of memory savings in mPnP-LLM, hence, depends on the LLM size and could be relatively smaller when LLM is small. See Table 1 and Table 2 in Section 4 of the paper for more details.

For answer generation, we adopt the prompt structure ``question:\{Q\}\texttt{</s>}answer:\{A\}\texttt{</s>}'', where ``\texttt{</s>}'' is the EOS token that has been already included in OPT and BLOOMZ's pre-trained embeddings. We truncate the question sequences so that the length of every preprocessed input sequence is within 64 tokens. On the testing data, we use a beam search size of 4, and set the maximum number of generated tokens to 8.

In all experiments, we conduct training in BF16 data format. We use a batch size of 16 and retrain the model for 8 epochs for modality adaptation. We set 8 epochs because no noticeable accuracy improvement is further observed when training more epochs. We use the AdamW optimizer [\citenum{loshchilov2017decoupled}] at a learning rate of $2\times10^{-5}$ with linear schedule and weight decay of $10^{-2}$. For mPnP-LLM's trainable connections, we set the learning rate to be $10^{-1}$. In each run, we use the \emph{exact match ratio} as the accuracy metric to indicate how many generated answers exactly match the ground truth, and measure PFLOPs as the training cost. Note that, sentence-level correctness is challenging when the QA task is open-ended text generation, compared to QA task of classification where the answers are categorized into several classes in advance. 

\section{Discussions}

In this section, we discuss various technical issues regarding mPnP-LLM's expandability and generalizability that the reviewers may be concerned of.

\vspace{0.05in}
\noindent\textbf{Extending to more modalities.} 
In this paper, we present the design of mPnP-LLM using an example of adapting between two input modalities for the multimodal QA task. However, we would like to highlight that, by design, mPnP-LLM can support adaptation among an arbitrary number of input modalities, since there is no constraint on the size of the K-V set in LLMs. On the other hand, simultaneously mounting more modalities could consume more computing resources on the embodied AI devices. In extreme cases, it would be necessary to meticulously rank the modalities' importance and only mount the most important modalities to fit to the devices' resource constraints. Quantifying such importances of different modalities at runtime will be our future work.
\vspace{0.05in}

\noindent\textbf{Generality on other generative AI tasks.} 
In this paper, we use the LLM as the backbone of mPnP-LLM for multimodal reasoning. However, the design rationale of mPnP-LLM can also be applied to other transformer-based generative backbones. For example, mPnP-LLM can also be applied to a pre-trained time-series transformer decoder [\citenum{zeng2023transformers}], for simpler embodied AI tasks such as motion planning. In this case, the projected multimodal K-V pairs can be concatenated with the previous tokens of motion status in the decoder blocks, to generate the future motions to perform. Even for non-transformer-based generative AI backbones such as stable diffusion models [\citenum{yang2022diffusion}], the design principle of mPnP-LLM is still applicable, as we can establish trainable latent connections with encoders. Efficient connections, in this case, will require specific model structure designs, which will be an interesting research direction to be explored in the future but is out of scope of this paper.
\vspace{0.05in}

\noindent\textbf{Training data for runtime modality adaptation.} mPnP-LLM requires labeled and modality-aligned training data for runtime modality adaptation. In practical embodied AI applications, it may be difficult to collect such training data at runtime, and we instead suggest preparing such multimodal training data in the offline phase. Such multimodal training data could include all the available input data modalities for every possible modality adaptation that may be necessary at runtime, and the offline preparation of such training data is consistent with our described workflow for modality adaptation in Section 3.4, where all the unimodal encoders for different input modalities and their pre-trained aligners are available offline.

In addition, as demonstrated by our experiment results in Section 4.5, mPnP-LLM is highly data efficient and only requires a few hundreds of training samples for effective modality adaptation. This data efficiency largely reduces the difficulty of preparing multimodal training data offline. 

\vspace{0.05in}


\noindent\textbf{Integration with encoder-decoder LLMs.}
Our design of mPnP-LLM focuses on decoder-only LLMs, which is currently the dominant LLM architecture and has been adopted by most existing LLMs, including the well-known GPT models. Encoder-decoder LLMs, on the other hand, can be considered as a sparse version of  decoder-only LLMs with the same parameter size, and they hence have weaker generative power. Nevertheless, mPnP-LLM can also be applied to the encoder-decoder LLM, by connecting the multimodal encoders to its decoder part. For encoder-decoder LLMs, by default, the K-V pairs in the decoder block's self-MHA are extended. The encoder part, on the other hand, functions as a separate encoding process over input text, and hence does not require any special handling.
\vspace{0.05in}

\noindent\textbf{Using a shared set vs. separate sets of connections for different input modalities.}
In this paper, our design of mPnP-LLM uses a shared set of trainable connections, including trainable weights of these connections, to all the involved input modalities. In this way, we can minimize the amount of connections and hence minimize the cost of runtime training. With that being said, the design of mPnP-LLM also allows using a separate set of connections for each modality's encoder to the LLM, so that each modality has more flexibility to interact with different levels of text representations. Such flexibility, of course, comes with higher training cost, and it is essential to balance between the flexibility and extra computing cost.
\vspace{0.05in}

\noindent\textbf{Runtime adaptation of the number of connections between input modality encoders and the LLM.}
In mPnP-LLM's design, the number of connections, which is also the number of last LLM blocks to be connected, is determined offline based on the user-defined computation budgets. Larger computation budgets allow more connections, which correspond to longer backpropagation paths in training and also higher costs of runtime training. Conceptually, such number of connections can also be flexibly adjusted at runtime, if the computing budgets vary in embodied AI tasks. More specifically, to reduce the number of connections at runtime, we can simply discard the existing connections and their corresponding trainable weights. Reversely, to add new connections at runtime, we can append new trainable weights between input modalities' encoders and the corresponding LLM blocks. No extra computing cost will be incurred for such runtime adaptation of the number of connections.
\vspace{0.05in}


\end{document}